\documentclass[11pt]{article}
\usepackage[margin=1in]{geometry}
\usepackage{times}
\usepackage{graphicx}
\usepackage{amsmath, amssymb}
\usepackage{booktabs}
\usepackage{multirow}
\usepackage{tabularx}
\usepackage{siunitx}
\usepackage{hyperref}
\usepackage[numbers,sort&compress]{natbib}
\usepackage{capt-of}
\usepackage{xcolor}
\usepackage{authblk}
\usepackage{tikz}
\usetikzlibrary{arrows.meta,positioning,calc}
\tikzset{>=Latex} 
\usepackage{float}
\usepackage{subcaption}

\hypersetup{
  colorlinks=true,
  linkcolor=blue,
  citecolor=blue,
  urlcolor=blue
}

\newcommand{\etal}{\textit{et al.}}

\newcommand{\eg}{\textit{e.g.}}

\title{A Unified Attention U-Net Framework for Cross-Modality Tumor Segmentation in MRI and CT}

\author[1]{Nishan Rai\thanks{Correspondence to: \texttt{nishan2@nmsu.edu}}}
\author[2]{Pushpa R. Dahal}

\affil[1]{Department of Computer Science, New Mexico State University}
\affil[2]{Department of Plant and Environmental Sciences, New Mexico State University}

\date{2025}

\begin{document}
\maketitle

\begin{abstract}
We study whether a single segmentation network can serve two imaging domains that differ in modality, anatomy, and lesion size. We train one Attention U-Net jointly on brain MRI (BraTS 2021) and lung CT (LIDC-IDRI), using a common input format, attention-gated skip connections, and a Focal Tversky loss, without modality-specific encoders or domain adaptation. On held-out test data, the model reaches a Dice score of 0.83 on MRI but only 0.55 on CT, while the aggregated score across both datasets is 0.64. The gap shows that a single aggregate metric can hide large differences between domains, and that unified models should be evaluated and reported per domain. We release our code as a simple, reproducible starting point for cross-modality tumor segmentation.
\end{abstract}
\section{Introduction}
Deep convolutional encoder--decoder networks such as U-Net~\citep{ronneberger2015unet} are the de facto standard for medical image segmentation.
Attention mechanisms~\citep{oktay2018attentionunet} and loss functions designed for class imbalance~\citep{salehi2017tversky,abraham2019focaltversky} have further improved sensitivity to small lesions. However, most work trains and evaluates models on a single modality and organ, whereas clinical practice routinely involves several modalities (\eg, MRI for the brain and CT for the lungs).
This raises a practical question: can a single model, trained jointly across modalities and anatomical sites, learn features that serve both, without modality-specific encoders or domain adaptation? We investigate this question by training a single Attention~U-Net on two widely used datasets: BraTS~2021 brain MRI~\citep{baid2021brats} and LIDC-IDRI lung CT~\citep{armato2011lidc}.
While cross-modality domain adaptation is well studied (\eg, SIFA~\citep{chen2020sifa}, Dou~\etal~\citep{dou2018crossmodality}), our emphasis is different: rather than adapting a trained model to a new modality, we examine whether a \emph{unified} model trained jointly on heterogeneous data can serve as a simple, resource-friendly baseline.
Our contributions are as follows: (1) a unified Attention U-Net trained jointly on MRI (BraTS) and CT (LIDC-IDRI) tumor datasets with balanced mixed batches; (2) a training recipe combining a brief CT warm-up, attention-gated skip connections, and a Focal Tversky loss; and (3) a per-domain evaluation showing that the model performs well on MRI but substantially worse on CT, a difference that an aggregated metric conceals.

\section{Related Work}

\textbf{U-Net and its Variants.}
Convolutional encoder--decoder networks have become the backbone of biomedical segmentation since the introduction of U-Net~\citep{ronneberger2015unet}.
Many variants have since been proposed.
nnU-Net~\citep{isensee2021nnunet} introduced a self-configuring framework that adapts preprocessing, architecture, and training automatically to a given dataset.
Nested architectures such as UNet++~\citep{zhou2018unetpp} refine skip connections with dense convolutional blocks to capture multi-scale features.
More recently, hybrid architectures combining convolution and transformers, including TransUNet~\citep{chen2021transunet} and Swin-UNet~\citep{cao2021swinunet}, model long-range dependencies and report strong results on medical segmentation benchmarks.

\textbf{Losses for Class Imbalance.}
Lesion segmentation is challenging due to the high class imbalance between small tumor regions and the background.
While Dice loss is commonly used, Salehi \etal~\citep{salehi2017tversky} introduced the Tversky loss as a generalization to control false positives and false negatives.
Abraham and Khan~\citep{abraham2019focaltversky} extended this with the Focal Tversky loss to emphasize harder-to-segment regions.
Other alternatives include boundary-aware losses such as Boundary Loss~\citep{kervadec2019boundary} and set-based objectives like the Lovász-Softmax loss~\citep{berman2018lovasz}, both of which aim to better align optimization with evaluation metrics such as IoU.

\textbf{Multi-Modal Segmentation.}
Several methods combine complementary information across imaging modalities.
HeMIS~\citep{havaei2016hemis} fused available modalities in a shared latent space, enabling inference when some modalities are missing, and Chartsias~\etal~\citep{chartsias2017hemis} learned modality-invariant latent representations for multimodal MR synthesis.
Valindria \etal~\citep{valindria2018multi} investigated unpaired multi-modal learning across CT and MRI for multi-organ segmentation, while Zhou~\etal~\citep{zhou2019multi} showed that self-supervised pretraining (Models Genesis) yields transferable representations for 3D medical image analysis.

\textbf{Cross-Modality Domain Adaptation.}
Another line of work focuses on mitigating domain shift between modalities by learning modality-invariant features.
Dou \etal~\citep{dou2018crossmodality} introduced adversarial learning for cross-modality segmentation.
Chen \etal~\citep{chen2020sifa} proposed SIFA, a synergistic framework combining image-level and feature-level adaptation between MRI and CT.
Similarly, Huo \etal~\citep{huo2018synsegnet} developed SynSegNet, which leverages synthetic images to bridge modality gaps.
These methods perform well but typically require more complex training pipelines, such as adversarial objectives or image translation.

\textbf{Our Contribution.}
In contrast to these more complex strategies, this work asks how far a single, unified model trained jointly on mixed batches of MRI and CT data can go. Rather than designing modality-specific encoders or domain adaptation schemes, we prioritize simplicity and report performance separately for each domain, so that the strengths and failures of the shared model remain visible.
\section{Datasets}
\label{sec:datasets}

\textbf{BraTS~2021 (MRI).}
We utilized the RSNA-ASNR-MICCAI BraTS~2021 challenge dataset~\citep{baid2021brats}, comprising pre-operative multi-parametric MRI scans (T1, T1ce, T2, FLAIR), illustrated in Figure 1a. We randomly sampled 200 patients and used every fourth axial slice. All tumor sub-regions were merged into a single binary whole-tumor mask. These modalities provide complementary soft tissue contrasts that facilitate accurate tumor delineation.

\textbf{LIDC-IDRI (CT).}
We employed the LIDC-IDRI thoracic CT database ~\citep{armato2011lidc}, consisting of lung CT scans annotated by four radiologists. A pixel was labeled as lesion if at least two of the four radiologists marked it (Figure 1b). Because LIDC-IDRI annotates lung nodules, not all of which are malignant, we use ``tumor'' in a broad sense to refer to annotated lesions in both datasets.

\begin{figure}[H]
\centering
\includegraphics[width=1\linewidth]{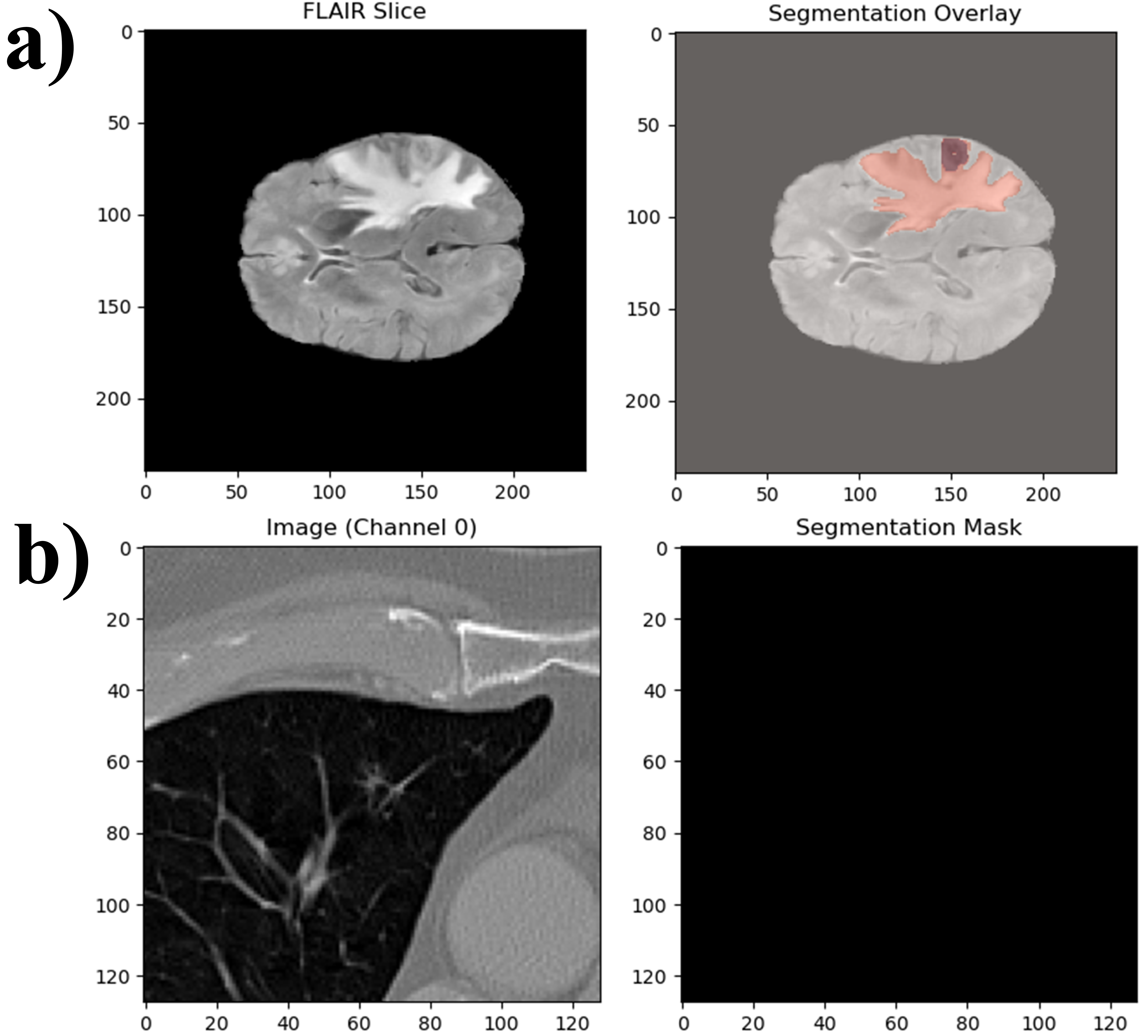}\hspace{0.04\linewidth}
\caption{Representative slices from the two datasets. (a) BraTS 2021 MRI slice (FLAIR) with the tumor mask overlaid. (b) LIDC-IDRI CT slice with the lung nodule mask.}
\label{fig:datasets}
\end{figure}

\section{Methods}
\subsection{Preprocessing and Augmentation}
To ensure consistent input dimensions across datasets, all axial slices were resized to $128 \times 128$ pixels. Each MRI slice was min--max scaled to $[0,1]$ per modality, and each CT slice was standardized to zero mean and unit variance and clipped to $[-3, 3]$. For the BraTS dataset, MRI volumes were sliced axially, and the four modalities (T1, T1ce, T2, FLAIR) were stacked to form 4-channel input tensors~\citep{baid2021brats}. In the case of the LIDC-IDRI CT slices~\citep{armato2011lidc}, which are natively grayscale, channel replication was employed to generate 4-channel tensors, ensuring architectural consistency with the MRI inputs.

To improve generalization, we applied geometric and photometric augmentations via the Albumentations library~\citep{buslaev2020albumentations}. These included random horizontal flips, random $90^\circ$ rotations, shift--scale--rotate transforms (rotations up to $\pm 15^\circ$), brightness and contrast adjustments, elastic and grid distortions, and coarse dropout. Augmentations were applied on the fly during training to simulate variation in anatomy and acquisition conditions.

\subsection{Network Architecture}
\label{sec:architecture}
We adopt the Attention U-Net architecture~\citep{oktay2018attentionunet}, which enhances the standard U-Net backbone~\citep{ronneberger2015unet} by integrating attention gates into the skip connections. This mechanism allows the model to selectively suppress irrelevant encoder features while highlighting salient regions in the decoder, thereby improving segmentation accuracy for irregular tumor boundaries.

\textbf{Backbone Structure.}
The network follows a symmetric encoder–decoder design. The encoder consists of two stages that progressively compress spatial resolution while expanding feature depth (channels: $4 \rightarrow 64 \rightarrow 128$). A bottleneck layer further expands the representation to 256 channels and applies dropout ($p=0.3$) for regularization. The decoder reconstructs spatial detail through two upsampling stages (channels: $256 \rightarrow 128 \rightarrow 64$). Skip connections bridge corresponding encoder and decoder layers at matching scales to preserve high-frequency spatial information.

\textbf{Attention Gates.}
At each skip connection, an attention gate filters encoder features $x$ based on a gating signal $g$ derived from the decoder. The attention coefficients are computed as:
\begin{equation}
    \alpha = \sigma \left( \psi^{T} \, \mathrm{ReLU} \left( W_x x + W_g g + b \right) \right),
    \label{eq:att_gate_new}
\end{equation}
where $W_x$, $W_g$, and $\psi$ are $1 \times 1$ convolutions followed by batch normalization, $b$ is a bias term, and $\sigma$ denotes the sigmoid activation. The modulated output is obtained via element-wise multiplication:
\begin{equation}
    x' = \alpha \cdot x,
\end{equation}
which suppresses background responses while propagating tumor-relevant features forward.

\begin{figure}[t]
\centering
\resizebox{\textwidth}{!}{%
\begin{tikzpicture}[font=\scriptsize, node distance=20mm and 10mm]
  \tikzstyle{inout} =[draw, rounded corners=2pt, minimum width=22mm, minimum height=9mm, align=center, fill=gray!20]
  \tikzstyle{enc}  =[draw, rounded corners=2pt, minimum width=22mm, minimum height=9mm, align=center, fill=orange!20]
  \tikzstyle{dec}  =[draw, rounded corners=2pt, minimum width=22mm, minimum height=9mm, align=center, fill=cyan!20]
  \tikzstyle{bot}  =[draw, rounded corners=2pt, minimum width=22mm, minimum height=9mm, align=center, fill=red!20]
  \tikzstyle{att}  =[draw, rounded corners=2pt, minimum width=22mm, minimum height=9mm, align=center, fill=green!20]
  \tikzstyle{flow} =[-{Latex[length=2mm]}, thick]
  \tikzstyle{skip} =[dashed, -{Latex[length=2mm]}, thick]

  \node[inout] (in) {Input \\ (4 ch)};
  \node[enc, right=of in] (e1) {Encoder 1 \\ (4 $\to$ 64)};
  \node[enc, right=of e1] (e2) {Encoder 2 \\ (64 $\to$ 128)};
  \node[bot, right=of e2] (bot) {Bottleneck \\ (128 $\to$ 256) \\ Dropout 0.3};

  \node[dec, right=of bot] (d1) {Decoder 1 \\ (256 $\to$ 128)};
  \node[dec, right=of d1] (d2) {Decoder 2 \\ (128 $\to$ 64)};
  \node[inout, right=of d2] (out) {Output \\ (1 ch)};

  \node[att, below=12mm of d1] (ag2) {Attention Gate 2 \\ (128)};
  \node[att, below=20mm of d2] (ag1) {Attention Gate 1 \\ (64)};

  \draw[flow] (in) -- (e1);
  \draw[flow] (e1) -- (e2);
  \draw[flow] (e2) -- (bot);
  \draw[flow] (bot) -- (d1);
  \draw[flow] (d1) -- (d2);
  \draw[flow] (d2) -- (out);

  \draw[flow] (ag2.north) -- (d1.south);
  \draw[flow] (ag1.north) -- (d2.south);

  \draw[skip] (e2.south) |- (ag2.west);
  \draw[skip] (e1.south) |- (ag1.west);

  \draw[skip] (e1.north) .. controls +(0,12mm) and +(0,12mm) .. (d2.north);
  \draw[skip] (e2.north) .. controls +(0,12mm) and +(0,12mm) .. (d1.north);

\end{tikzpicture}%
}
\vspace{2mm}
\begin{tikzpicture}[font=\scriptsize]
  \node[draw, align=left, fill=white, rounded corners=2pt, inner sep=4pt] {
    \textbf{Legend:}\quad
    \fcolorbox{black}{gray!20}{\rule{4mm}{3mm}} Input/Output\quad
    \fcolorbox{black}{orange!20}{\rule{4mm}{3mm}} Encoder\quad
    \fcolorbox{black}{cyan!20}{\rule{4mm}{3mm}} Decoder\quad
    \fcolorbox{black}{red!20}{\rule{4mm}{3mm}} Bottleneck\quad
    \fcolorbox{black}{green!20}{\rule{4mm}{3mm}} Attention Gate\quad
  };
\end{tikzpicture}

\caption{Architectural schematic of the proposed Attention U-Net model. The Input/Output blocks (gray) define the data ingestion and final segmentation layers. The Encoder (orange) functions as the contracting path, compressing spatial dimensions to extract hierarchical features. The Bottleneck (red) captures high-level semantic representations, incorporating dropout for regularization. The Decoder (blue) serves as an expanding path, upsampling feature maps to reconstruct the segmentation mask. Attention Gates (green) selectively filter encoder feature maps prior to concatenation to suppress irrelevant activations, while Skip Connections (dashed lines) preserve critical spatial details.}
\label{fig:arch}
\end{figure}
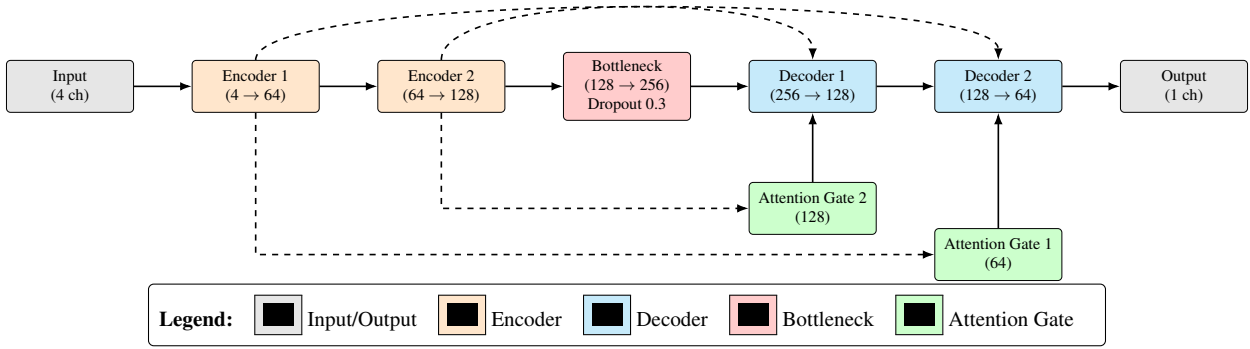

\textbf{Segmentation Head.}
A final $1 \times 1$ convolution maps decoder features to a single-channel probability map, followed by a sigmoid activation to produce the binary segmentation mask. The full network comprises approximately 1.89 million trainable parameters and requires $\sim$7.3\,G multiply–add operations, providing a lightweight yet expressive backbone suitable for joint MRI+CT training. Architectural details are illustrated in Fig.~\ref{fig:arch}.

\subsection{Loss Function and Optimization }

To address class imbalance between tumor and background pixels, we employ the Focal Tversky Loss~\citep{abraham2019focaltversky}. This loss extends the Tversky Index by introducing a focal parameter that down-weights easily classified examples, encouraging the model to focus on hard-to-segment regions.

Let $p_i \in [0,1]$ denote the predicted probability for pixel $i$, and let $g_i \in \{0,1\}$ denote the corresponding ground-truth label. The Tversky Index (TI) is defined as:
\begin{equation}
TI(p, g) = 
\frac{\sum_i p_i g_i}
{\sum_i p_i g_i 
+ \alpha \sum_i (1 - p_i) g_i 
+ \beta \sum_i p_i (1 - g_i)},
\label{eq:tversky_new}
\end{equation}
where $\alpha$ and $\beta$ control the penalties for false negatives and false positives, respectively.

The corresponding Tversky loss is:
\begin{equation}
\mathcal{L}_{\text{Tversky}} = 1 - TI(p, g).
\end{equation}

To further emphasize hard examples, the Focal Tversky loss introduces a focusing parameter $\gamma$:
\begin{equation}
\mathcal{L}_{\text{FTL}} = \left( 1 - TI(p, g) \right)^{\gamma},
\label{eq:focal_tversky_new}
\end{equation}
which increases the contribution of poorly segmented examples to the total loss. We use $\gamma = 0.75$.

Network parameters are optimized using the AdamW optimizer~\citep{loshchilov2019adamw}. 
To accelerate convergence, we adopt the 1-cycle learning rate policy~\citep{smith2018superconvergence}, which linearly increases the learning rate to a maximum value and subsequently decreases it during the remainder of training. 
All experiments are implemented in PyTorch~\citep{paszke2019pytorch}.

\subsection{Unified Training Strategy}

We adopt a single Attention U-Net that is shared across both imaging modalities, without modality-specific encoders or task-specific output heads. To encourage stable convergence on CT, we first pretrain the network for 5 epochs on LIDC-IDRI alone with Adam (learning rate $1\times10^{-4}$) and the Focal Tversky loss with $(\alpha, \beta) = (0.5, 0.5)$. The resulting weights are then used to initialize the unified model.

Subsequently, we fine-tune the network jointly on BraTS and LIDC-IDRI using balanced minibatches. Each training batch is constructed to contain an equal number of slices from each dataset (a 1:1 ratio), so that the optimizer receives comparable signal from MRI and CT domains. The training loss is a weighted sum ($0.4$ and $0.6$) of two Focal Tversky terms with $(\alpha, \beta) = (0.8, 0.2)$ and $(0.5, 0.5)$, each computed on one half of the batch. We optimize with AdamW (base learning rate $1\times10^{-4}$, peak $3\times10^{-4}$ under the one-cycle policy, weight decay $1\times10^{-5}$), clip gradients to a norm of 1.0, and use a batch size of 8. Training runs for up to 45 epochs with early stopping (patience of 3 epochs). All experiments use a fixed random seed of 42.

Each dataset is randomly split into training (70\%), validation (15\%), and test (15\%) sets at the slice level. Validation is performed separately on the held-out BraTS and LIDC-IDRI splits. At each epoch, we compute validation Dice and loss per modality and select the best checkpoint based on the macro-averaged Dice across MRI and CT to avoid overfitting to a single domain.

\section{Experiments}
\label{sec:experiments}

\subsection{Evaluation Protocol}

Segmentation performance is assessed using standard metrics, including the Dice Similarity Coefficient (Dice), Intersection over Union (IoU), Precision, Recall, F1-score, and the Area Under the Receiver Operating Characteristic Curve (ROC-AUC). Predicted probabilities are binarized at a threshold of 0.1, and Dice, IoU, precision, recall, and F1 are computed for each test slice and averaged. ROC-AUC is computed from pixel-level probabilities pooled over the test set.

Let $p$ denote the predicted binary segmentation mask and $g$ the corresponding ground-truth mask. True positives (TP), false positives (FP), true negatives (TN), and false negatives (FN) are computed on a pixel-wise basis.

\begin{align}
\text{Dice}(p,g) &= \frac{2 |p \cap g|}{|p| + |g|}, \\
\text{IoU}(p,g) &= \frac{|p \cap g|}{|p \cup g|}, \\
\text{Precision} &= \frac{TP}{TP + FP}, \quad
\text{Recall} &= \frac{TP}{TP + FN}, \\
F1 &= \frac{2 \cdot \text{Precision} \cdot \text{Recall}}
           {\text{Precision} + \text{Recall}}.
\end{align}

\subsection{Quantitative Results}
Table~\ref{tab:metrics} reports test performance for the unified model. On MRI (BraTS), the model reaches a Dice score of 0.83 and an AUC of 0.97. On CT (LIDC-IDRI), performance is substantially lower, with a Dice score of 0.55 and an AUC of 0.83. The unified row, computed over the combined test sets (Dice 0.64, AUC 0.89), lies between the two and, on its own, would obscure this gap.

\begin{table}[h]
\centering
\caption{Test performance metrics by dataset (unified model).}
\label{tab:metrics}
\begin{tabular}{lcccccc}
\toprule
Dataset & Dice & IoU & Precision & Recall & F1 & AUC \\
\midrule
BraTS   & 0.83 & 0.80 & 0.86 & 0.92 & 0.84 & 0.97 \\
LIDC    & 0.55 & 0.49 & 0.74 & 0.69 & 0.57 & 0.83 \\
Unified & 0.64 & 0.59 & 0.78 & 0.77 & 0.66 & 0.89 \\
\bottomrule
\end{tabular}
\end{table}

\subsection{Training Dynamics}
\begin{figure}[H]
\centering
\includegraphics[width=1\linewidth]{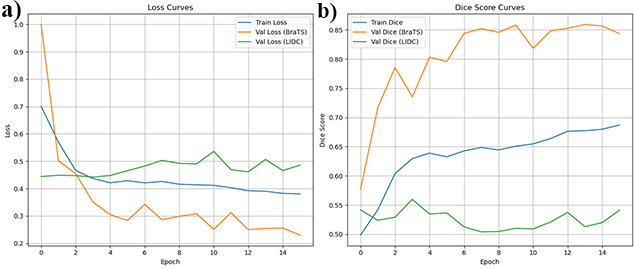}
\caption{Training and validation curves showing a. loss and b. Dice score across epochs for both MRI (BraTS) and CT (LIDC-IDRI).}
\label{fig:training_curves}
\end{figure}

The training and validation trajectories for loss and Dice score are depicted in Figure~\ref{fig:training_curves}. A rapid decline in training and validation loss is observed for the BraTS dataset, accompanied by a corresponding increase in the Dice score, which reaches approximately 0.8 by epoch 5. Conversely, the LIDC validation loss exhibits higher volatility and remains elevated throughout training, consistent with the challenges posed by smaller lesion sizes and higher noise levels in CT data. Despite this, the Dice score for LIDC shows gradual improvement, stabilizing around 0.55. These dynamics indicate that the unified model learns MRI segmentation quickly, while CT remains considerably harder for the shared network.

\subsection{ROC and Confusion Analysis}

\begin{figure}[H]
\centering
\includegraphics[width=1\linewidth]{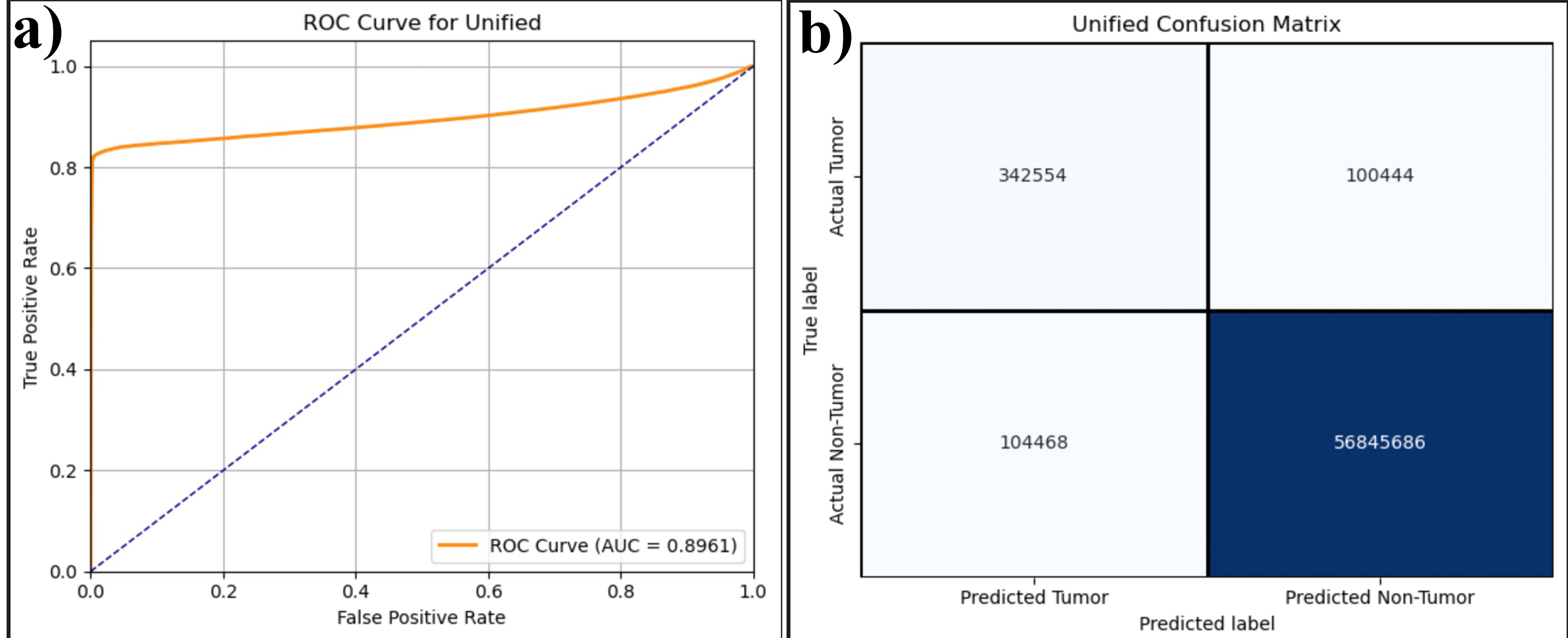}\hspace{0.04\linewidth}
\caption{Performance analysis of the unified Attention U-Net framework on both datasets. 
(a) Unified ROC curve for test set (AUC = 0.89). 
(b) Unified confusion matrix.}
\label{fig:roc_confusion}
\end{figure}

To further analyze decision thresholds, Fig.~\ref{fig:roc_confusion}a shows the unified ROC curve over both test sets (AUC = 0.89), indicating that the model ranks tumor pixels above background reasonably well even where its thresholded masks are less accurate.
Together with the per-dataset recall in Table~\ref{tab:metrics} (0.92 on MRI versus 0.69 on CT), the confusion matrix (Fig.~\ref{fig:roc_confusion}b) indicates that missed tumor pixels account for much of the error on CT.

\subsection{Qualitative Results}
\begin{figure}[H]
\centering
\includegraphics[width=1\linewidth]{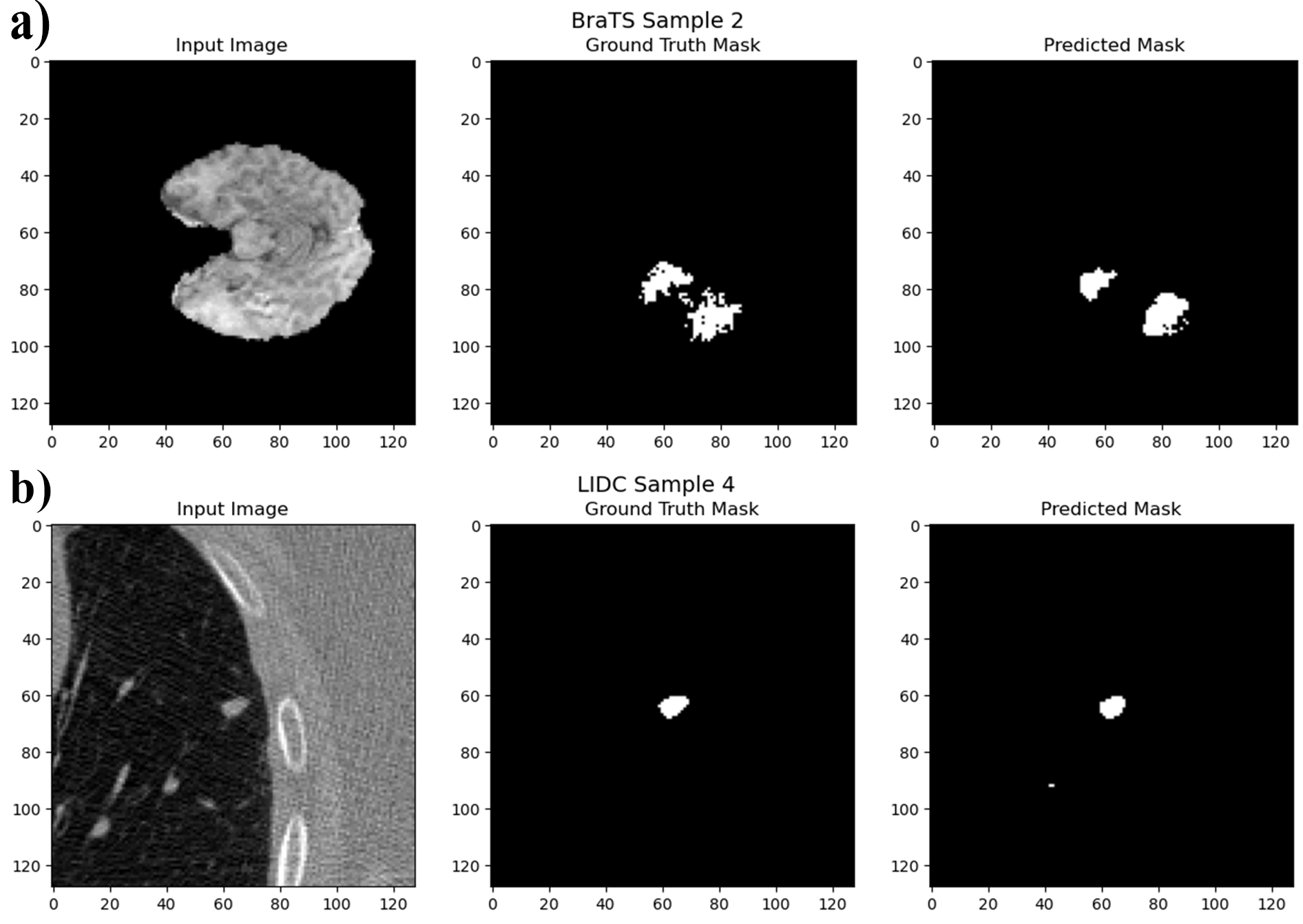}\hspace{0.04\linewidth}
\caption{Qualitative segmentation results for the unified model. The top panel a. displays results for BraTS Sample 2, and the bottom panel b. for LIDC Sample 4. Each sample row presents the input image, ground truth mask, and predicted mask to visualize segmentation performance on MRI and CT modalities.}
\label{fig:examples}
\end{figure}

Representative segmentation results for MRI and CT slices are displayed in Figure~\ref{fig:examples}. On MRI, predicted masks follow tumor boundaries closely. On CT, the model sometimes under-segments small nodules, consistent with its lower recall on LIDC-IDRI. 

\section{Discussion}
The unified Attention U-Net performs well on MRI (BraTS), with high recall, and moderately on CT (LIDC-IDRI), where precision (0.74) exceeds recall (0.69). The most informative result is the gap itself: an aggregated Dice of 0.64 would suggest a moderately effective model, while the per-domain scores show a model that works well on one task and struggles on the other. Several factors may contribute, including the small size of many lung nodules relative to the $128 \times 128$ input resolution, differences in annotation protocols, and the intensity and contrast shift between MRI and CT. Our experiments do not isolate these factors. 

Qualitatively, the shared network handles both domains without modality-specific components. In BraTS Sample~2 (Fig.~\ref{fig:examples}a), predictions align closely with complex tumor boundaries and edema despite substantial variation in MRI intensity, and in LIDC Sample~4 (Fig.~\ref{fig:examples}b), the model segments a lung nodule embedded within heterogeneous thoracic tissue. These examples are consistent with attention gates helping to suppress irrelevant structures such as the skull or ribs, although we did not test this directly by comparing against a model without attention.

Compared with domain adaptation approaches that rely on adversarial training or image translation, the unified model is simpler to train and deploy. The intensity and contrast shift between MRI and CT nevertheless remains a central challenge. Lightweight modality-aware adapters or domain conditioning may improve CT performance while preserving this simplicity.

\section{Limitations and Societal Impact}
This study has several limitations. First, we do not compare against separately trained single-modality models or a standard U-Net, and we do not ablate the attention gates or the Focal Tversky loss, so we cannot attribute the MRI--CT gap to joint training or to any single design choice. Second, the data were split at the slice level rather than the patient level, so slices from the same patient can appear in both training and test sets; patient-level splits would give a stricter estimate of generalization. Third, we do not report variance across repeated runs or confidence intervals. Fourth, experiments were run on a single GPU without extensive hyperparameter sweeps, and resizing slices to $128 \times 128$ may disadvantage small lung nodules. Finally, treating 3D volumes as independent 2D slices discards spatial context; volumetric architectures are a natural next step.

Any clinical use of such a model would require validation across diverse patient populations, scanners, and institutions, since performance may drop on data that differ from the training distribution. Because segmentation errors could delay diagnosis, a system of this kind should serve only as decision support for radiologists, with human oversight, and not as an autonomous diagnostic tool.

\section{Conclusion}
We trained a single Attention U-Net jointly on brain MRI (BraTS 2021) and lung CT (LIDC-IDRI) without modality-specific encoders or domain adaptation. The shared model segments MRI tumors well (Dice 0.83) but performs substantially worse on CT nodules (Dice 0.55), and an aggregated metric (Dice 0.64) hides this difference. We therefore recommend that unified multi-domain models be evaluated and reported per domain. Natural next steps include single-modality baselines, ablations of the attention gates and loss, volumetric architectures, and methods that let a model recognize when an input falls outside the conditions it handles well.

\section{Acknowledgements}
This work began as the final project for Applied Machine Learning I (CS 519) at New Mexico State University. The authors thank the Department of Computer Science at New Mexico State University for facilities and technical support. Furthermore, the authors acknowledge Assistant Prof. Tuan Le for the opportunity to conduct this research and for his guidance throughout the project.

\paragraph{Reproducibility and Code Availability}
To facilitate reproducibility, all preprocessing scripts, training schedules, random seeds, and configuration defaults have been released publicly via our project repository: \url{https://github.com/Nishan8912/unified-attention-unet-cross-modality}.

\bibliographystyle{unsrtnat}
\bibliography{references}
\end{document}